\documentclass[sigconf, 9pt, nonacm]{acmart}
\usepackage{balance}
\usepackage{multirow}
\graphicspath{{./images/}}
\newcommand{\NA}{\textemdash}
\newcommand{\meanstd}[2]{\ensuremath{#1{\scriptstyle\,\pm\,#2}}}

\ccsdesc[500]{Computing methodologies~Neural networks}
\ccsdesc[300]{Cross-computing tools and techniques~Design}
\ccsdesc[300]{Computer systems organization~Embedded and cyber-physical systems}
\setcopyright{none}
\acmConference[NeuroPHY~'26]{Neuromorphic Physical Layer Signal Processing for Wireless Systems Workshop}{2026}{co-located with EWSN}
\renewcommand\footnotetextcopyrightpermission[1]{}

\begin{document}

\title{Predictive Suppression Layers for Communication-Efficient Spiking Neural Networks}

\author{Aidin Attar}
\affiliation{%
  \institution{University of Padova}
  \department{Department of Information Engineering}
  \city{Padova}
  \country{Italy}}
\email{aidin.attar@phd.unipd.it}

\author{Michele Rossi}
\affiliation{%
  \institution{University of Padova}
  \department{Dept. of Inf. Eng. and Dept. of Mathematics}
  \city{Padova}
  \country{Italy}}
\email{michele.rossi@unipd.it}

\renewcommand{\shortauthors}{Attar and Rossi}

\begin{abstract}
Feedforward Spiking Neural Networks (SNNs) typically propagate every generated spike indiscriminately, disregarding whether the information is redundant from an information-theoretic perspective. This lack of selectivity induces high redundancy in inter-layer communication, creating an expensive overhead, e.g., in scenarios involving many-core neuromorphic hardware or communication-dominated Internet-of-Things (IoT) where features are transmitted wirelessly. To address this challenge, we trade localized processing for leaner network channels by introducing a minimal predictive coding framework for SNNs. We propose two layer variants sharing a predictor block: {\it error units}, which transmit signed spiking residuals, and {\it predictive suppression}, which uses residual magnitude to dynamically gate and forward only unpredictable, ``surprising'' activity. Evaluated on the N-MNIST and Spiking Heidelberg Digits (SHD) datasets using diagnostic metrics that decouple local processing from cross-layer communication, our new predictive coding layers achieve significant communication savings. Numerical results reveal a three-fold reduction in communicated activity, while increasing the task accuracy for both datasets. The latter finding is notable, and suggests that predictive coding layers not only minimize communication overhead, but also produce output feature vectors with a higher representation power.
\end{abstract}

\keywords{spiking neural networks, predictive coding, communication efficiency,
  neuromorphic computing, event-based sensing}

\maketitle

\section{Introduction}
\label{sec:intro}

Spiking neural networks are a class of artificial neural networks that closely mimic the behavior of the brain. Their constituting neurons exchange information through sparse, discrete spikes instead of dense activations, mapping naturally onto neuromorphic hardware where activity, not idle time, entails energy consumption. This makes the \emph{content} and \emph{timing} of in-network communication central to their design. Yet, in a standard feedforward SNN every spike a layer produces is propagated forward indiscriminately, with no notion of whether that spike carries new information or merely repeats what the next layer could already have anticipated. The total spike count metric, commonly used to gauge energy efficiency, does not capture this distinction. Total spike count aggregates two qualitatively different costs: the \emph{local} processing a layer performs internally, and the activity it \emph{communicates} to downstream layers. On hardware where inter-core communication dominates the energy budget, these costs {\it are not interchangeable}, and a metric that conflates them may be inapt to come up with efficient designs. This same asymmetry appears in signal-processing and transmission in edge-systems: moving spike activity between devices is often {\it substantially more expensive} than performing local processing on a given device. In IoT, communication accounts for most of the energy consumed by IoT nodes and usually dominates local processing~\cite{lopez-ardao_current_2021}.
Under such communication-dominated regimes, we seek a mechanism that at the cost of an increased local processing leads to a leaner channel.

\begin{figure}[htb]
\centering
\includegraphics[width=\columnwidth]{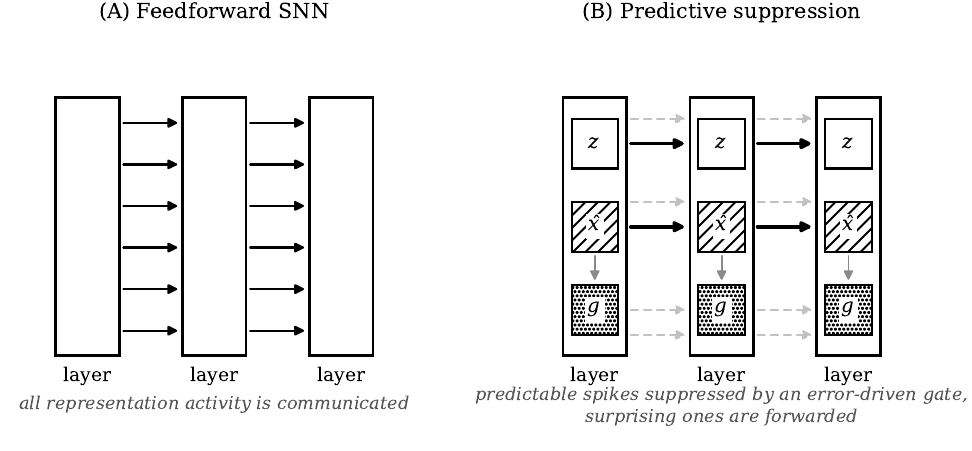}
\vspace{-20pt}
\caption{Predictive coding layer. (A) Plain feedforward SNN forwards the entire representation to the next layer. (B) Predictive suppression keeps the layers spiking but adds a local error-driven gate that suppresses predictable spikes and forwards the surprising ones, giving less communicated activity.}
\Description{Schematic comparison between a plain feedforward spiking network, where every spike a layer produces is forwarded to the next layer, and predictive suppression, where local error-driven gates suppress predictable spikes and let surprising ones through.}
\label{fig:intro}
\end{figure}

Predictive coding offers a useful design principle to control the local processing {\it vs} communication tradeoff. In its classical form, a cortical area transmits the part of its input that a higher area cannot predict, in place of the full input~\citep{rao_predictive_1999,friston_free-energy_2010}. Interpreted as a communication rule, this suggests forwarding ``surprise'' rather than the original signal. Translating this principle into an SNN, however, comes at a price. In fact, to know what is predictable, a layer must be in the position of computing a prediction, and this entails an increase in its local processing. Predictive coding therefore does not reduce the total number of spikes but rather increases it. The relevant question is whether local predictive computation (the price to pay) can be leveraged for a {\it more selective and informative} inter-layer (or inter-device, e.g., in an IoT setting) channel. In this paper, we propose minimal predictive coding as the mechanism to control this tradeoff. According to it, each layer locally predicts its own input and uses the resulting prediction error to gate the latent spikes it sends to the next layer. As a result of this gating mechanism, predictable spikes are suppressed, by only letting those carrying new information through. Fig.~\ref{fig:intro} contrasts the proposed predictive design with a plain feedforward baseline.

Specifically, we propose two minimal predictive coding variants. The first one, referred to as ``error unit'', forwards prediction errors in place of the original latent vector produced by a layer. The second variant, referred to as ``predictive suppression'', modulates the output latent vector via a gating mechanism driven by the error magnitude. Our numerical results confirm that both approaches increase the total spike count of a layer. However, the number of spikes that are communicated to the next layer decreases substantially, leading to a leaner channel. On N-MNIST~\citep{orchard_converting_2015}, error unit layers improve classification accuracy over a matched feedforward baseline by concentrating communication resources on high-error (unpredictable) inputs. Predictive suppression provides a slightly higher performance, still beating a non-predictive SNN baseline. A layer-wise linear-probe analysis shows that its compressed message remains as decodable as the full representation; a large share of the removable redundancy appears at the first communicated layer, which carries about $7\times$ less weighted activity than the SNN baseline. On the more challenging SHD~\citep{cramer_heidelberg_2022} dataset, predictive suppression layers still shine, by improving over the matched SNN baseline, whereas error units underperform. In general, predictive suppression is found to be the best design, delivering more stable performance across the considered datasets.

To summarize, in this work we introduce two minimal predictive coding layer designs for SNNs, performing a communication-oriented evaluation that separates local layer processing from inter-layer communication costs. The proposed predictive layers are evaluated on N-MNIST and SHD datasets, demonstrating their effectiveness in delivering surprise-selective, low-redundancy communication channels while retaining competitive task accuracy and signal reconstruction performance at a decoder. Typical results provide a three-fold reduction in communication activity, while the task accuracy increases for both datasets, suggesting that our designs not only minimize communication overhead, but also produce output feature vectors with a higher representation power. 

\section{Related work}
\label{sec:related}

Predictive coding was originally proposed as a theory of cortical processing where higher areas predict lower-level activity and only the unpredicted residual is propagated forward~\citep{rao_predictive_1999}. Rao and Ballard~\citep{rao_predictive_1999} used this principle to account for extra-classical receptive-field effects in the visual cortex, framing prediction errors as the informative signals exchanged between hierarchical areas. Friston~\citep{friston_free-energy_2010} later embedded the same intuition in the free-energy principle, where perception and action are cast as processes that minimise prediction error under a generative model. More recently, predictive-coding networks have been studied as machine-learning models. Whittington and Bogacz~\citep{whittington_approximation_2017} showed that local predictive-coding dynamics can approximate error backpropagation under suitable assumptions, while Millidge et al.~\citep{millidge_predictive_2022} reviewed predictive coding as a possible route toward learning systems that are less dependent on backpropagation. In the present work we take a narrower interpretation: prediction error is used as a communication signal that gates what a spiking layer forwards, without involving iterative top-down inference or a new learning rule. 

Deep spiking neural networks are commonly trained with surrogate gradients, which replace the non-differentiable derivative of the spike function with a smooth proxy during backpropagation. \textit{SuperSpike} introduced one such surrogate-gradient formulation for multilayer SNNs~\citep{zenke_superspike_2018}, and Neftci et al.~\citep{neftci_surrogate_2019} surveyed how surrogate gradients make gradient-based optimisation practical for spiking models. Bellec et al.~\citep{bellec_solution_2020} addressed a related credit-assignment problem in recurrent SNNs with eligibility propagation, showing how local eligibility traces can support learning in temporal spiking networks. These works are primarily concerned with biologically plausible learning and credit assignment in spiking networks. Here, we use the supervised surrogate-gradient setting pragmatically, with the goal of designing and evaluating a communication-efficient predictive layer for edge-oriented SNNs.

A separate line of work studies efficiency in SNNs and neuromorphic systems. Neuromorphic processors such as INTEL's Loihi are designed to exploit sparse event-driven activity and on-chip communication~\citep{davies_loihi_2018}, motivating models that reduce spike count, enforce sparse activity, or exploit conditional execution. More broadly, conditional-computation methods, such as adaptive computation time~\citep{graves_adaptive_2016}, learn to allocate computation selectively instead of applying the same processing everywhere. These works establish the importance of spiking activity and communication costs, but typically {\it report a single aggregate quantity} such as total spike count or executed operations. In contrast, we separate local predictive activity from inter-layer communication: a layer can spend more activity locally to predict its input while still reducing or restructuring the signal it sends forward.

Predictive coding in spiking networks is an active but fragmented line of work. Ororbia~\citep{ororbia_spiking_2023} proposed spiking neural predictive coding for continual learning from data streams, using predictive objectives to support adaptation in spiking models. N'dri et al.~\citep{ndri_predictive_2026} recently surveyed predictive coding with SNNs, covering models where spiking dynamics, local errors, and predictive objectives are combined in biologically inspired learning systems. Difference Predictive Coding takes a more communication-oriented view, by transmitting sparse differences between predictions and inputs when training SNNs~\citep{karlsson_difference_2026}. These works show that prediction errors can be represented or used in spiking systems, but they do not investigate whether a layer can trade local predictive computation for a more selective inter-layer message while preserving task information.

The closest prior art is Predictive Coding Light (PCL)~\citep{ndri_predictive_2025}, which builds a hierarchical spiking network where predictable spikes are suppressed and a compressed representation is forwarded, very close in spirit to our predictive-suppression mechanism. PCL is unsupervised, relies on spike-timing-dependent inhibitory plasticity, and is framed as a biologically plausible representation-learning model. Differently, our model is supervised and trained with surrogate gradients, the gate is explicitly driven by the prediction error, and the analysis is centred on communication: total activity is decomposed into local processing and communicated parts, weighted and event-based communication are kept separate, surprise selectivity and redundant communication are measured, and the communicated message is tested for linear decodability and task accuracy.

\section{Methods}
\label{sec:method}

\begin{figure*}[t]
\centering
\includegraphics[width=0.81\textwidth]{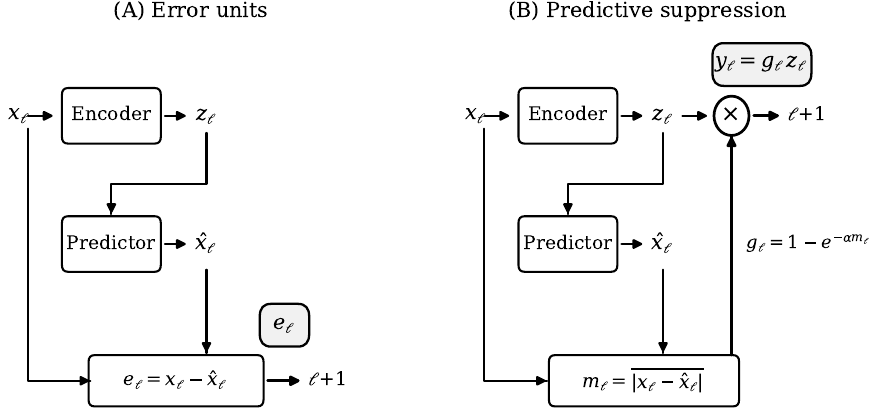}
\vspace{-10pt}
\caption{What each layer communicates. The two schemes share the same encoder/predictor skeleton; only the forwarded message differs. (A) The error-units variant forwards the prediction error $e_\ell=x_\ell-\hat{x}_\ell$. (B) Predictive suppression encodes $z_\ell$, predicts its own input $\hat{x}_\ell$, and forwards $y_\ell=g_\ell z_\ell$, where the gate $g_\ell=1-e^{-\alpha m_\ell}$ is driven by the prediction-error magnitude.}
\Description{Two-panel block diagram of the per-layer mechanisms. Both panels show a shared skeleton with an encoder, a predictor, the prediction estimate, and the layer input routed into a bottom box. In the left panel (error units) the bottom box computes the prediction error and forwards it to the next layer. In the right panel (predictive suppression) the bottom box computes the prediction-error magnitude that drives a multiplicative gate on the encoder output, and the gated representation is forwarded.}
\label{fig:concept}
\end{figure*}

\subsection{Predictive layer designs}
\label{sec:predictive_layers}
Both predictive variants use the same local encoder-predictor skeleton, see Fig.~\ref{fig:concept}. For an input $x_\ell[t]$ at layer $\ell$ and time step $t$, the layer computes a spiking latent representation and a local reconstruction:
\begin{align}
z_\ell[t] &= \mathrm{Enc}_\ell(x_\ell[t]), &
\hat{x}_\ell[t] &= \mathrm{Pred}_\ell(z_\ell[t]),
\label{eq:enc_pred}
\end{align}
where $\mathrm{Enc}_\ell$ is a linear-leaky-integrate-and-fire (LIF) encoder and $\mathrm{Pred}_\ell$ is a shallow LIF-based predictor mapping $z_\ell[t]$ back to the layer input dimensionality. The prediction residual and its scalar magnitude are
\begin{align}
e_\ell[t] &= x_\ell[t] - \hat{x}_\ell[t], &
m_\ell[t] &= \frac{1}{D}\sum_{d=1}^{D} \left|e_\ell[t,d]\right|.
\label{eq:error_magnitude}
\end{align}
The two variants differ in what is communicated to the next layer.

\noindent \paragraph{Error units.}
The error-units variant follows the direct predictive-coding interpretation and forwards the residual
\begin{equation}
y_\ell[t] = \mathrm{LIF}\!\left([\,\mathrm{relu}(e_\ell[t]);\;\mathrm{relu}(-e_\ell[t])\,]\right),
\label{eq:error_units}
\end{equation}
in order to take into account negative errors.

\noindent \paragraph{Predictive suppression.}
In this case, the residual is used as a control signal, whose magnitude defines an error-driven gate,
\begin{equation}
g_\ell[t] =
\mathrm{clamp}\!\left(1-e^{-\alpha m_\ell[t]},\, g_{\min},\, 1\right),
\label{eq:gate}
\end{equation}
where $\mathrm{clamp}$ bounds the gate between $g_{\min}$ and $1$. Hence, the layer communicates a gated version of its latent spike train:
\begin{equation}
y_\ell[t] = g_\ell[t] z_\ell[t].
\label{eq:graded}
\end{equation}
When the input is predicted well, the gate closes toward $g_{\min}$ and predictable spikes are suppressed; when the input is surprising, the gate opens and spikes are forwarded. We set $\alpha=10$ after a preliminary sweep over $\alpha\in\{5,7,10\}$, which gave stable accuracy, and use $g_{\min}=0.05$ as a small floor to avoid fully closing the communication channel.

The representation $z_\ell[t]$ is binary, while the gate $g_\ell[t]$ is continuous. Thus the communicated signal in Eq.~\eqref{eq:graded} is {\it graded}: a predictive amplitude modulated spike train, not a pure binary event stream. The receiving layer remains spiking, integrating $y_\ell[t]$ and emitting its own spikes. Training minimises a task loss, a predictive (reconstruction) loss, and a min-rate regulariser that keeps the predictor from collapsing to silence,
\begin{equation}
\mathcal{L} = \mathcal{L}_{\text{task}} + \beta\big(\mathcal{L}_{\text{pred}} + \lambda\,\mathcal{L}_{\text{rate}}\big),
\label{eq:loss}
\end{equation}
with $\beta=0.1$ and $\lambda=2$, where $\mathcal{L}_{\text{rate}}=[\,0.05-\bar{r}_{\text{pred}}\,]_+^2$ penalises predictor firing rates that fall below a target rate of $0.05$.

\subsection{Strictly event-driven predictive suppression}
\label{sec:variants}
Predictive suppression in ~\eqref{eq:graded} uses a graded inter-layer channel. Here, we additionally define two binary variants that retain the same error-driven gate, while enforcing strictly event-based communication. A hard threshold with a straight-through estimator, termed \textbf{hard-STE}, emits $y_\ell[t]=\mathbf{1}[g_\ell[t]>\theta]\,z_\ell[t]$ in the forward pass, with gradients flowing through the underlying continuous gate. The second variant, termed \textbf{respike}, re-binarises the graded current through a LIF neuron, $y_\ell[t]=\mathrm{LIF}(\gamma\,g_\ell[t]\,z_\ell[t])$, yielding a more conventional SNN block. The hard-threshold is empirically set to $\theta=0.3$ on N-MNIST and $\theta=0.1$ on SHD; the respike gain is $\gamma=4$.

\section{Evaluation metrics}
\label{sec:metrics}

Our methodological approach is to separate where activity is spent between local processing and communication, instead of using a single total spike count. The total activity cost is thus
\begin{equation}
E_{\text{total}} = E_{\text{local}} + E_{\text{comm}},
\end{equation}
where $E_{\text{local}}$ accounts for local processing activity (encoding, prediction), whereas  $E_{\text{comm}}$ is the activity communicated across layers. The unweighted sum is the special case where local and communicated activities have equal cost. To account for communication-dominated regimes, we consider the weighted proxy
\begin{equation}
C(\rho) = E_{\text{local}} + \rho\,E_{\text{comm}},
\label{eq:weighted-cost}
\end{equation}
where $\rho$ represents the relative cost of communicated activity; $\rho{=}1$ recovers the unweighted total, while $\rho{>}1$ models regimes in which transmitting activity is more expensive than local computing. $C(\rho)$ is an activity-proxy sensitivity: it allows to gauge whether the local predictive cost is compensated for by the reduction in inter-layer communication. For the communicated part, two measures characterize the communicated activity:
\begin{equation}
E_{\text{comm}}^{\text{w}} = \!\sum_{t,d} g_\ell[t]\,z_\ell[t,d],
\quad
E_{\text{comm}}^{\text{ev}} = \!\sum_{t,d} \mathbf{1}[y_\ell[t,d]\neq 0].
\end{equation}
The weighted measure on the left captures the modulated activity of the proposed graded model. The event measure on the right counts how many binary spikes are actually transmitted, the natural quantity for the binary variants and the SNN baseline.

Beyond the activity decomposition we use three diagnostic metrics. Let \(E_{\mathrm{repr},\ell}\) be the activity of the local spiking representation (before suppression) at layer $\ell$ and $E_{\mathrm{comm},\ell}$ the activity forwarded to the next layer. The \emph{compression ratio}
\[
\mathrm{Comp}_\ell = \frac{E_{\mathrm{comm},\ell}}{E_{\mathrm{repr},\ell}}
\]
measures how much of the local representation at layer $\ell$ is communicated to the next layer; lower values indicate stronger suppression.

Compression alone says nothing about whether communication is suppressed selectively. We define two error-conditioned diagnostics. Within each batch, the $(t,b)$ time-data batch pairs of a layer are ranked by their local prediction-error magnitude $m_\ell[t]$, and the top and bottom quartiles ($25\%$) form the high- and low-error subsets, respectively. If \(C_{\mathrm{high}}\) and \(C_{\mathrm{low}}\) denote the mean communicated activity on these two subsets, the \emph{surprise-selectivity index} (SSI) is defined as 
\[
\mathrm{SSI}
=
\frac{C_{\mathrm{high}} - C_{\mathrm{low}}}
     {C_{\mathrm{high}} + C_{\mathrm{low}}}.
\]
Specifically, a high SSI (ideally approaching $1$) means that communication activity across layers concentrates on surprising inputs, i.e., that $C_{\mathrm{high}}$ dominates $C_{\mathrm{low}}$. The companion \emph{redundant-communication fraction} (RCF)
\[
\mathrm{RCF}
=
\frac{C_{\mathrm{low,total}}}
     {C_{\mathrm{total}}}
\]
measures the share of total communication spent on low-error inputs; low RCF means little activity is wasted on predictable inputs. Linear probes trained on the time-averaged communicated signal and, separately, on the full local representation check whether the compressed message remains linearly decodable into the class label.

\section{Experimental setup}
\label{sec:setup}

Testing is performed on two event-based benchmarks that are compatible with a temporal multilayer architecture: N-MNIST~\citep{orchard_converting_2015} ($2{\times}34{\times}34$, ten classes) and Spiking Heidelberg Digits~\citep{cramer_heidelberg_2022} ($700$ channels, twenty classes). Events are binned into $T=16$ frames. All models are multilayer spiking networks with matched hidden sizes ($256$-$192$-$128$) and identical preprocessing; the matched baseline is a plain feedforward LIF network of the same capacity. Training uses surrogate gradients via SpikingJelly~\citep{fang_spikingjelly_2023} and the Adam optimiser, with five seeds. Preliminary sweeps over $\alpha\in\{5,7,10\}$ left N-MNIST accuracy stable ($0.975$-$0.976$, Table~\ref{tab:main}); we use $\alpha=10$ as the main operating point. All reported cost metrics are activity proxies, not measured hardware energy.

\section{Numerical Results}
\label{sec:results}

\subsection{Predictive coding trades local layer processing for selective communication}
\label{sec:results-nmnist}

\begin{figure*}[t]
\centering
\includegraphics[width=0.82\textwidth]{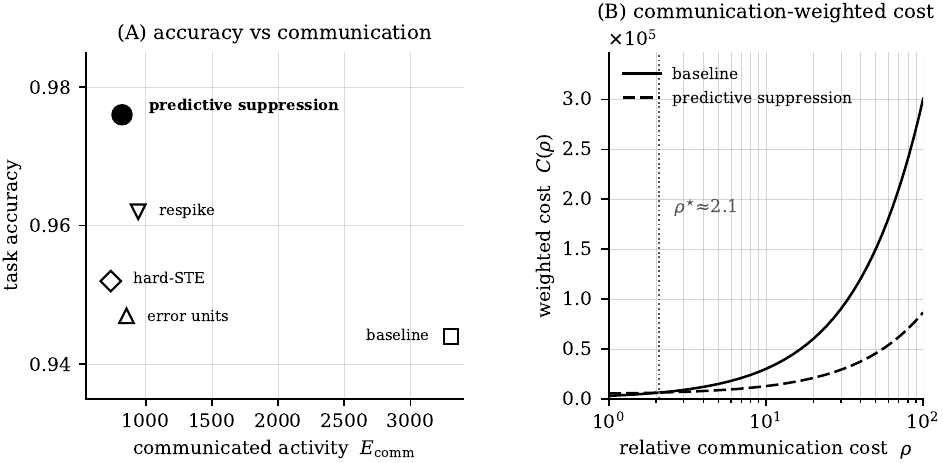}
\vspace{-10pt}
\caption{(A) Accuracy versus communicated activity proxy (summed over layers) on N-MNIST. Predictive suppression (filled circle) keeps high accuracy while communicating less than the matched baseline. (B) Communication-weighted cost sensitivity $C(\rho)=E_{\mathrm{local}}+\rho\,E_{\mathrm{comm}}$. As predictive suppression spends extra local processing to reduce communication, it becomes cheaper than the baseline once communication is weighted about $\rho^{\star}\!\approx\!2.1$ times more than local computation.}
\Description{Two-panel figure on N-MNIST. The left panel is a scatter plot of model accuracy versus weighted communicated activity, with predictive suppression sitting at the top-left (high accuracy, low communication) and the matched baseline at the bottom-right. The right panel is a log-log plot of the weighted cost C(rho) as a function of the relative communication cost rho between 1 and 1000, with two curves: a solid line for the baseline and a dashed line for predictive suppression; the dashed line is above the solid one at low rho and crosses below it at rho about 2.1, marked by a vertical dotted line.}
\label{fig:accvscomm}
\end{figure*}

Our results confirm that, on N-MNIST suppressing predictable spikes does not reduce task accuracy. In fact, the proposed predictive coding schemes improve over the matched SNN baseline while using less aggregate weighted inter-layer communication, see Table~\ref{tab:main}. As expected, the predictor adds local computation within a layer, but the signal sent between layers becomes leaner and more task-relevant. Fig.~\ref{fig:accvscomm} summarises this trade-off and shows how it changes as communication becomes more expensive than local processing (increasing $\rho$).

\begin{table*}[t]
\centering
\caption{Summary of main results. Accuracy is mean $\pm$ standard deviation over five seeds. SSI and RCF are only reported for suppression-based communication. $E_{\mathrm{comm}}$ is computed on the actual transmitted signal $y_\ell$; it is a summed magnitude for graded channels and an event count for binary channels.}
\label{tab:main}
\vspace{-10pt}
\small
\setlength{\tabcolsep}{5pt}
\begin{tabular}{lllcccc}
\toprule
Dataset & Model & Channel & Accuracy & $E_{\text{comm}}$ & Compr. & SSI / RCF \\
\midrule
\multirow{6}{*}{N-MNIST}
 & SNN baseline                           & events   & \meanstd{0.941}{0.002} & 3006 & 1.00 & \NA \\
 & Error units                            & events   & \meanstd{0.946}{0.004} & 855  & 0.33 & \NA \\
 & Predictive suppression & weighted & \meanstd{0.976}{0.001} & 821  & 0.52 & 0.770 / 0.050 \\
 & Respike                                & events   & \meanstd{0.959}{0.002} & 943  & 0.81 & 0.995 / 0.001 \\
 & Hard-STE                               & events   & \meanstd{0.950}{0.012} & 736  & 0.73 & 0.985 / 0.003 \\
\midrule
\multirow{5}{*}{SHD}
 & SNN baseline                           & events   & \meanstd{0.621}{0.005} & 3094 & 1.00 & \NA \\
 & Error units                            & events   & \meanstd{0.271}{0.012} & 151  & 0.17 & \NA \\
 & Predictive suppression          & weighted & \meanstd{0.717}{0.004} & 747  & 0.51 & 0.920 / 0.020 \\
 & Hard-STE                               & events   & \meanstd{0.698}{0.009} & 1119 & 0.89 & 0.930 / 0.020 \\
 & Respike                                & events   & \meanstd{0.639}{0.009} & 931  & 0.75 & 0.950 / 0.013 \\
\bottomrule
\end{tabular}
\end{table*}

\begin{figure}[htbp]
\centering
\includegraphics[width=\columnwidth]{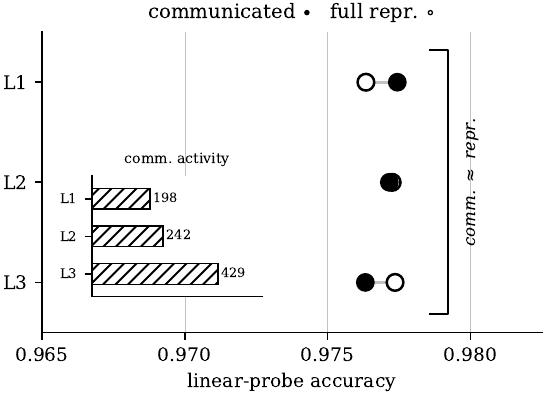}
\vspace{-15pt}
\caption{Linear-probe decodability for predictive suppression. At every layer the communicated message (filled) is essentially as decodable as the full representation (open); the paired dots nearly coincide. The first layer message carries about $7\times$ less weighted activity than the baseline. Inset: the communicated activity that this message costs, per layer.}
\Description{Dumbbell plot of linear-probe accuracy at three layers on N-MNIST. Each layer is a horizontal row with two markers, a filled dot for the probe trained on the communicated message and an open dot for the probe trained on the full local representation; the two markers nearly overlap at every layer. A small inset shows the communicated activity used per layer.}
\label{fig:probe}
\end{figure}

\paragraph{Decodability of the communicated message.}
A compact channel is useful only if it preserves task information. To verify this, we train linear probes on two signal types at each layer: the compressed message communicated to the next layer, and the full local spiking representation before suppression. Across layers, the two probe accuracies nearly coincide, see Fig.~\ref{fig:probe}. This demonstrates that predictive coding removes activity that the downstream decoder does not need, while preserving a linearly accessible class signal. At the first layer the prediction-driven gate removes the largest share of redundancy: the message carries $7\times$ less weighted activity than the baseline ($198$ versus $1397$) while remaining as linearly decodable as the full representation.

\begin{figure}[htbp]
\centering
\includegraphics[width=\columnwidth]{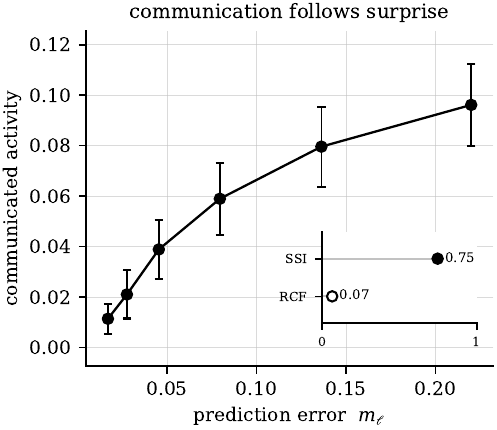}
\vspace{-15pt}
\caption{Communication is selective, not merely reduced: at the first layer of predictive suppression, communicated activity rises monotonically with the pred.-error magnitude $m_\ell$. Inset: SSI (filled) is high and the RCF (open) is low.}
\Description{Plot of mean communicated activity at the first predictive-suppression layer on N-MNIST as a function of the per-layer prediction-error magnitude, with error bars; the curve rises monotonically. A small inset shows two short horizontal lollipops: a filled marker for the surprise-selectivity index near 0.75 and an open marker for the redundant-communication fraction near 0.07.}
\label{fig:selectivity}
\end{figure}

\noindent \paragraph{Prediction-error selectivity.}
Decodability shows that the message is sufficient, it does not show that the gate is prediction-driven. In fact, a random gate could preserve class information while still wasting activity on predictable inputs.  Fig.~\ref{fig:selectivity} shows that this is not the case: the communicated activity monotonically increases with the prediction-error magnitude, the SSI is high and the RCF fraction is small. As expected from the error-driven gate, communication increases with prediction error: ``surprising'' inputs receive more communication activity, while predictable ones are strongly suppressed.

\subsection{Event-driven variants and temporal transfer}
\label{sec:results-variants}

Predictive suppression layers use a graded inter-layer channel because the gate is continuous. We now test whether the same gate can support strictly binary communication.

As shown in Table~\ref{tab:main}, training binary messages end-to-end recovers much of the graded model's performance, but the preferred binarization method depends on the dataset. On N-MNIST, \textbf{respike} is the best performing binary variant. On SHD, \textbf{hard-STE} is the preferred method, although it has the weakest compression. Predictive suppression, i.e., the graded channel option, is the most robust design overall.

\paragraph{Transfer to SHD}
SHD provides a temporal test beyond event-based vision. Under the same simple feedforward architecture, predictive suppression again improves over the matched SNN baseline, with high selectivity and low redundant communication (Table~\ref{tab:main}). Instead, the error-units variant performs poorly in this setting. This suggests that the error is more useful as a control signal for communication than as the message itself.

\section{Discussion}
\label{sec:discussion}

Predictive coding increases the total spiking activity via the addition of a local predictor. This allows changing  what is sent between layers, attaining a compact, {\it surprise-oriented} and linearly decodable channel. This type of efficiency departs from spike-count minimisation and becomes relevant when communication is more expensive than local layer processing, as on hardware where moving activity between cores dominates the energy budget or IoT edge nodes where output features are to be transmitted over a wireless channel.

The cost curve in Fig.~\ref{fig:accvscomm}B plots the weighted cost $C(\rho)$ (see Eq.~\eqref{eq:weighted-cost}) for a predictive suppression layer and a standard SNN layer as a function of factor $\rho$, gauging how much communication activity costs with respect to local processing. As can be seen, predictive suppression becomes convenient as it crosses the SNN baseline at $\rho^{\star} \approx 2.1$. In such regimes, predictive suppression becomes cheaper than the SNN baseline.  

The continuous predictive suppression gate trains reliably and yields the most compact output channel. Its event-driven variants show that the approach can be pushed to binary, genuinely spike-based communication, although which variant works best depends on the dataset. Event-driven predictive suppression is feasible, but which binary mechanism is best is not universal.

\section{Conclusions}
\label{sec:conclusion}

In this paper we proposed novel {\it predictive coding} designs for SNNs. The new layer type integrates a predictor that is utilized to infer and forward only non-predictable input to downstream layers. The rationale is that encoding ``surprise'' should lead to a leaner channel between layers, minimizing the amount of communicated activity. This is particularly appealing for exchanging features across processing cores or for edge IoT devices where SNN features are to be communicated via a wireless channel to a decision point. Our numerical results, obtained on N-MNIST and SHD datasets confirm the effectiveness of the proposed approach, namely, the channel becomes surprise-oriented and low-redundancy. Moreover, output features lead to better accuracies than with the matched SNN baseline and are more linearly decodable. Top-down and iterative extensions of the predictive suppression layers, convolutional gates to handle spatial event data, and direct hardware-energy measurements are natural extensions to our work.

\begin{acks}
The work in this paper is supported by the Multi-X project that has received funding from the Smart Networks and Services Joint Undertaking under the European Union’s Horizon Europe R\&I programme under GA No. 101192521.
\end{acks}

\balance
\bibliographystyle{ACM-Reference-Format}
\bibliography{references}


\begin{thebibliography}{17}


\ifx \showCODEN    \undefined \def \showCODEN     #1{\unskip}     \fi
\ifx \showISBNx    \undefined \def \showISBNx     #1{\unskip}     \fi
\ifx \showISBNxiii \undefined \def \showISBNxiii  #1{\unskip}     \fi
\ifx \showISSN     \undefined \def \showISSN      #1{\unskip}     \fi
\ifx \showLCCN     \undefined \def \showLCCN      #1{\unskip}     \fi
\ifx \shownote     \undefined \def \shownote      #1{#1}          \fi
\ifx \showarticletitle \undefined \def \showarticletitle #1{#1}   \fi
\ifx \showURL      \undefined \def \showURL       {\relax}        \fi
\providecommand\bibfield[2]{#2}
\providecommand\bibinfo[2]{#2}
\providecommand\natexlab[1]{#1}
\providecommand\showeprint[2][]{arXiv:#2}

\bibitem[Bellec et~al\mbox{.}(2020)]%
        {bellec_solution_2020}
\bibfield{author}{\bibinfo{person}{Guillaume Bellec}, \bibinfo{person}{Franz Scherr}, \bibinfo{person}{Anand Subramoney}, \bibinfo{person}{Elias Hajek}, \bibinfo{person}{Darjan Salaj}, \bibinfo{person}{Robert Legenstein}, {and} \bibinfo{person}{Wolfgang Maass}.} \bibinfo{year}{2020}\natexlab{}.
\newblock \showarticletitle{A solution to the learning dilemma for recurrent networks of spiking neurons}.
\newblock \bibinfo{journal}{\emph{Nature Communications}} \bibinfo{volume}{11}, \bibinfo{number}{1} (\bibinfo{date}{July} \bibinfo{year}{2020}), \bibinfo{pages}{3625}.
\newblock
\showISSN{2041-1723}


\bibitem[Cramer et~al\mbox{.}(2022)]%
        {cramer_heidelberg_2022}
\bibfield{author}{\bibinfo{person}{Benjamin Cramer}, \bibinfo{person}{Yannik Stradmann}, \bibinfo{person}{Johannes Schemmel}, {and} \bibinfo{person}{Friedemann Zenke}.} \bibinfo{year}{2022}\natexlab{}.
\newblock \showarticletitle{The {Heidelberg} {Spiking} {Data} {Sets} for the {Systematic} {Evaluation} of {Spiking} {Neural} {Networks}}.
\newblock \bibinfo{journal}{\emph{IEEE Transactions on Neural Networks and Learning Systems}} \bibinfo{volume}{33}, \bibinfo{number}{7} (\bibinfo{date}{July} \bibinfo{year}{2022}), \bibinfo{pages}{2744--2757}.
\newblock
\showISSN{2162-2388}


\bibitem[Davies et~al\mbox{.}(2018)]%
        {davies_loihi_2018}
\bibfield{author}{\bibinfo{person}{Mike Davies}, \bibinfo{person}{Narayan Srinivasa}, \bibinfo{person}{Tsung-Han Lin}, \bibinfo{person}{Gautham Chinya}, \bibinfo{person}{Yongqiang Cao}, \bibinfo{person}{Sri~Harsha Choday}, \bibinfo{person}{Georgios Dimou}, \bibinfo{person}{Prasad Joshi}, \bibinfo{person}{Nabil Imam}, \bibinfo{person}{Shweta Jain}, \bibinfo{person}{Yuyun Liao}, \bibinfo{person}{Chit-Kwan Lin}, \bibinfo{person}{Andrew Lines}, \bibinfo{person}{Ruokun Liu}, \bibinfo{person}{Deepak Mathaikutty}, \bibinfo{person}{Steven McCoy}, \bibinfo{person}{Arnab Paul}, \bibinfo{person}{Jonathan Tse}, \bibinfo{person}{Guruguhanathan Venkataramanan}, \bibinfo{person}{Yi-Hsin Weng}, \bibinfo{person}{Andreas Wild}, \bibinfo{person}{Yoonseok Yang}, {and} \bibinfo{person}{Hong Wang}.} \bibinfo{year}{2018}\natexlab{}.
\newblock \showarticletitle{Loihi: {A} {Neuromorphic} {Manycore} {Processor} with {On}-{Chip} {Learning}}.
\newblock \bibinfo{journal}{\emph{IEEE Micro}} \bibinfo{volume}{38}, \bibinfo{number}{1} (\bibinfo{date}{Jan.} \bibinfo{year}{2018}), \bibinfo{pages}{82--99}.
\newblock
\showISSN{1937-4143}


\bibitem[Fang et~al\mbox{.}(2023)]%
        {fang_spikingjelly_2023}
\bibfield{author}{\bibinfo{person}{Wei Fang}, \bibinfo{person}{Yanqi Chen}, \bibinfo{person}{Jianhao Ding}, \bibinfo{person}{Zhaofei Yu}, \bibinfo{person}{Timothée Masquelier}, \bibinfo{person}{Ding Chen}, \bibinfo{person}{Liwei Huang}, \bibinfo{person}{Huihui Zhou}, \bibinfo{person}{Guoqi Li}, {and} \bibinfo{person}{Yonghong Tian}.} \bibinfo{year}{2023}\natexlab{}.
\newblock \showarticletitle{{SpikingJelly}: {An} open-source machine learning infrastructure platform for spike-based intelligence}.
\newblock \bibinfo{journal}{\emph{Science Advances}} \bibinfo{volume}{9}, \bibinfo{number}{40} (\bibinfo{date}{Oct.} \bibinfo{year}{2023}), \bibinfo{pages}{1--70}.
\newblock


\bibitem[Friston(2010)]%
        {friston_free-energy_2010}
\bibfield{author}{\bibinfo{person}{Karl Friston}.} \bibinfo{year}{2010}\natexlab{}.
\newblock \showarticletitle{The free-energy principle: a unified brain theory?}
\newblock \bibinfo{journal}{\emph{Nature Reviews Neuroscience}} \bibinfo{volume}{11}, \bibinfo{number}{2} (\bibinfo{date}{Feb.} \bibinfo{year}{2010}), \bibinfo{pages}{127--138}.
\newblock
\showISSN{1471-0048}


\bibitem[Graves(2016)]%
        {graves_adaptive_2016}
\bibfield{author}{\bibinfo{person}{Alex Graves}.} \bibinfo{year}{2016}\natexlab{}.
\newblock \bibinfo{title}{Adaptive {Computation} {Time} for {Recurrent} {Neural} {Networks}}.
\newblock


\bibitem[Karlsson et~al\mbox{.}(2026)]%
        {karlsson_difference_2026}
\bibfield{author}{\bibinfo{person}{Ville Karlsson}, \bibinfo{person}{Nicklas Fianda}, {and} \bibinfo{person}{Joni-Kristian Kämäräinen}.} \bibinfo{year}{2026}\natexlab{}.
\newblock \showarticletitle{Difference {Predictive} {Coding} for {Training} {Spiking} {Neural} {Networks}}. In \bibinfo{booktitle}{\emph{Proceedings of the {International} {Conference} on {Learning} {Representations} ({ICLR}-26)}}. \bibinfo{address}{Rio de Janeiro, Brazil}, \bibinfo{pages}{1--20}.
\newblock


\bibitem[López-Ardao et~al\mbox{.}(2021)]%
        {lopez-ardao_current_2021}
\bibfield{author}{\bibinfo{person}{J.~Carlos López-Ardao}, \bibinfo{person}{Raúl~F. Rodríguez-Rubio}, \bibinfo{person}{Andrés Suárez-González}, \bibinfo{person}{Miguel Rodríguez-Pérez}, {and} \bibinfo{person}{M.~Estrella Sousa-Vieira}.} \bibinfo{year}{2021}\natexlab{}.
\newblock \showarticletitle{Current {Trends} on {Green} {Wireless} {Sensor} {Networks}}.
\newblock \bibinfo{journal}{\emph{Sensors}} \bibinfo{volume}{21}, \bibinfo{number}{13} (\bibinfo{date}{Jan.} \bibinfo{year}{2021}), \bibinfo{pages}{4281}.
\newblock
\showISSN{1424-8220}


\bibitem[Millidge et~al\mbox{.}(2022)]%
        {millidge_predictive_2022}
\bibfield{author}{\bibinfo{person}{Beren Millidge}, \bibinfo{person}{Tommaso Salvatori}, \bibinfo{person}{Yuhang Song}, \bibinfo{person}{Rafal Bogacz}, {and} \bibinfo{person}{Thomas Lukasiewicz}.} \bibinfo{year}{2022}\natexlab{}.
\newblock \showarticletitle{Predictive {Coding}: {Towards} a {Future} of {Deep} {Learning} beyond {Backpropagation}?}. In \bibinfo{booktitle}{\emph{Proceedings of the {Thirty}-{First} {International} {Joint} {Conference} on {Artificial} {Intelligence} ({IJCAI}-22)}}. \bibinfo{publisher}{International Joint Conferences on Artificial Intelligence}, \bibinfo{address}{Vienna, Austria}, \bibinfo{pages}{5538--5545}.
\newblock


\bibitem[Neftci et~al\mbox{.}(2019)]%
        {neftci_surrogate_2019}
\bibfield{author}{\bibinfo{person}{Emre~O. Neftci}, \bibinfo{person}{Hesham Mostafa}, {and} \bibinfo{person}{Friedemann Zenke}.} \bibinfo{year}{2019}\natexlab{}.
\newblock \showarticletitle{Surrogate {Gradient} {Learning} in {Spiking} {Neural} {Networks}: {Bringing} the {Power} of {Gradient}-{Based} {Optimization} to {Spiking} {Neural} {Networks}}.
\newblock \bibinfo{journal}{\emph{IEEE Signal Processing Magazine}} \bibinfo{volume}{36}, \bibinfo{number}{6} (\bibinfo{date}{Nov.} \bibinfo{year}{2019}), \bibinfo{pages}{51--63}.
\newblock
\showISSN{1558-0792}


\bibitem[N’dri et~al\mbox{.}(2025)]%
        {ndri_predictive_2025}
\bibfield{author}{\bibinfo{person}{Antony~W. N’dri}, \bibinfo{person}{Thomas Barbier}, \bibinfo{person}{Céline Teulière}, {and} \bibinfo{person}{Jochen Triesch}.} \bibinfo{year}{2025}\natexlab{}.
\newblock \showarticletitle{Predictive {Coding} {Light}}.
\newblock \bibinfo{journal}{\emph{Nature Communications}} \bibinfo{volume}{16}, \bibinfo{number}{1} (\bibinfo{date}{Oct.} \bibinfo{year}{2025}), \bibinfo{pages}{8880}.
\newblock
\showISSN{2041-1723}


\bibitem[N’dri et~al\mbox{.}(2026)]%
        {ndri_predictive_2026}
\bibfield{author}{\bibinfo{person}{Antony~W. N’dri}, \bibinfo{person}{William Gebhardt}, \bibinfo{person}{Céline Teulière}, \bibinfo{person}{Fleur Zeldenrust}, \bibinfo{person}{Rajesh P.~N. Rao}, \bibinfo{person}{Jochen Triesch}, {and} \bibinfo{person}{Alexander Ororbia}.} \bibinfo{year}{2026}\natexlab{}.
\newblock \showarticletitle{Predictive coding with spiking neural networks: {A} survey}.
\newblock \bibinfo{journal}{\emph{Neural Networks}}  \bibinfo{volume}{196} (\bibinfo{date}{April} \bibinfo{year}{2026}), \bibinfo{pages}{108371}.
\newblock
\showISSN{0893-6080}


\bibitem[Orchard et~al\mbox{.}(2015)]%
        {orchard_converting_2015}
\bibfield{author}{\bibinfo{person}{Garrick Orchard}, \bibinfo{person}{Ajinkya Jayawant}, \bibinfo{person}{Gregory~K. Cohen}, {and} \bibinfo{person}{Nitish Thakor}.} \bibinfo{year}{2015}\natexlab{}.
\newblock \showarticletitle{Converting {Static} {Image} {Datasets} to {Spiking} {Neuromorphic} {Datasets} {Using} {Saccades}}.
\newblock \bibinfo{journal}{\emph{Frontiers in Neuroscience}}  \bibinfo{volume}{9} (\bibinfo{date}{Nov.} \bibinfo{year}{2015}), \bibinfo{pages}{437}.
\newblock
\showISSN{1662-453X}


\bibitem[Ororbia(2023)]%
        {ororbia_spiking_2023}
\bibfield{author}{\bibinfo{person}{Alexander Ororbia}.} \bibinfo{year}{2023}\natexlab{}.
\newblock \showarticletitle{Spiking neural predictive coding for continually learning from data streams}.
\newblock \bibinfo{journal}{\emph{Neurocomputing}}  \bibinfo{volume}{544} (\bibinfo{date}{Aug.} \bibinfo{year}{2023}), \bibinfo{pages}{126292}.
\newblock
\showISSN{0925-2312}


\bibitem[Rao and Ballard(1999)]%
        {rao_predictive_1999}
\bibfield{author}{\bibinfo{person}{Rajesh P.~N. Rao} {and} \bibinfo{person}{Dana~H. Ballard}.} \bibinfo{year}{1999}\natexlab{}.
\newblock \showarticletitle{Predictive coding in the visual cortex: a functional interpretation of some extra-classical receptive-field effects}.
\newblock \bibinfo{journal}{\emph{Nature Neuroscience}} \bibinfo{volume}{2}, \bibinfo{number}{1} (\bibinfo{date}{Jan.} \bibinfo{year}{1999}), \bibinfo{pages}{79--87}.
\newblock
\showISSN{1546-1726}


\bibitem[Whittington and Bogacz(2017)]%
        {whittington_approximation_2017}
\bibfield{author}{\bibinfo{person}{James C.~R. Whittington} {and} \bibinfo{person}{Rafal Bogacz}.} \bibinfo{year}{2017}\natexlab{}.
\newblock \showarticletitle{An {Approximation} of the {Error} {Backpropagation} {Algorithm} in a {Predictive} {Coding} {Network} with {Local} {Hebbian} {Synaptic} {Plasticity}}.
\newblock \bibinfo{journal}{\emph{Neural Computation}} \bibinfo{volume}{29}, \bibinfo{number}{5} (\bibinfo{date}{May} \bibinfo{year}{2017}), \bibinfo{pages}{1229--1262}.
\newblock
\showISSN{0899-7667}


\bibitem[Zenke and Ganguli(2018)]%
        {zenke_superspike_2018}
\bibfield{author}{\bibinfo{person}{Friedemann Zenke} {and} \bibinfo{person}{Surya Ganguli}.} \bibinfo{year}{2018}\natexlab{}.
\newblock \showarticletitle{{SuperSpike}: {Supervised} {Learning} in {Multilayer} {Spiking} {Neural} {Networks}}.
\newblock \bibinfo{journal}{\emph{Neural Computation}} \bibinfo{volume}{30}, \bibinfo{number}{6} (\bibinfo{date}{June} \bibinfo{year}{2018}), \bibinfo{pages}{1514--1541}.
\newblock
\showISSN{0899-7667}


\end{thebibliography}

\end{document}